\pdfoutput=1

\documentclass[11pt]{article}

\usepackage[final]{acl}

\usepackage{times}
\usepackage{latexsym}
\usepackage[table]{xcolor}
\usepackage{amsmath}
\usepackage[T1]{fontenc}
\usepackage[utf8]{inputenc}

\usepackage{microtype}

\usepackage{inconsolata}
\usepackage{xcolor}
\usepackage{hyperref}       %
\usepackage{url}            %
\usepackage{booktabs}       %
\usepackage{amsfonts}       %
\usepackage{subcaption}
\usepackage{caption}

\usepackage{graphicx}
\usepackage{cleveref}

\usepackage{graphicx}
\usepackage{xcolor}
\usepackage{scalefnt}
\usepackage{adjustbox}

\crefformat{section}{\S#2#1#3}
\Crefname{equation}{Eq.}{Eqs.}
\Crefname{figure}{Fig.}{Figs.}
\Crefname{table}{Tab.}{Tabs.}
\Crefname{appendix}{\S$\!$}{\S$\!$}

\definecolor{col}{HTML}{3B82F6} %
\definecolor{mate}{HTML}{10B981} %

\title{
\raisebox{0.1em}{
\begin{adjustbox}{valign=c}
    \includegraphics[height=1.6em]{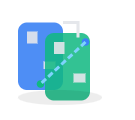}
\end{adjustbox}}
\hspace{-1.1em}
\textsc{\textbf{
\textcolor{col}{Col}\textcolor{mate}{Mate}
}}:  Contrastive Late Interaction and Masked Text \\ for Multimodal Document Retrieval
}

\newcommand{\colmate}[1]{\textsc{
\textcolor{col}{Col}\textcolor{mate}{Mate}}}

\author{
Ahmed Masry$^{\clubsuit\heartsuit}$ , Megh Thakkar$^{\clubsuit}$, Patrice Bechard $^{\clubsuit}$, \\\bf Sathwik Tejaswi Madhusudhan$^{\clubsuit}$, \bf Rabiul Awal$^{\clubsuit\spadesuit\diamondsuit}$, Shambhavi Mishra$^{\clubsuit\triangle}$, \\ \bf Akshay Kalkunte Suresh$^{\clubsuit}$, \bf Srivatsava Daruru$^{\clubsuit}$, 
Enamul Hoque$^{\heartsuit}$, \\ \bf Spandana Gella$^{\clubsuit}$, \bf Torsten Scholak$^{\clubsuit}$, Sai Rajeswar$^{\clubsuit\spadesuit\diamondsuit}$\thanks{Correpondance to ahmed.masry@servicenow.com and sai.rajeswar@servicenow.com.} \\ \\
$^\clubsuit$ServiceNow, 
$^\heartsuit$York University, 
$^\spadesuit$MILA - Quebec AI Institute, \\
$^\diamondsuit$Université de Montréal, 
$^\triangle$École de technologie supérieure \\
}

\begin{document}
\maketitle
\begin{abstract}
Retrieval-augmented generation has proven practical when models require specialized knowledge or access to the latest data. 
However, existing methods for multimodal document retrieval often replicate techniques developed for text-only retrieval, whether in how they encode documents, define training objectives, or compute similarity scores.
To address these limitations, we present\colmate{}, a document retrieval model that bridges the gap between multimodal representation learning and document retrieval.\colmate{} utilizes a novel OCR-based pretraining objective, a self-supervised masked contrastive learning objective, and a late interaction scoring mechanism more relevant to multimodal document structures and visual characteristics.\colmate{} obtains 3.61\% improvements over existing retrieval models on the ViDoRe V2 benchmark, demonstrating stronger generalization to out-of-domain benchmarks.
\end{abstract}

\section{Introduction}
\label{intro}

The information assimilated by LLMs (Large Language Models) during pretraining and stored in their parametric memory can get outdated with time~\cite{zhang2024evaluatingexternalparametricknowledge}. 
LLMs also face challenges when tackling tasks requiring highly-specialized knowledge or scenarios requiring information not present in their training data, such as confidential information~\cite{gao2024retrievalaugmentedgenerationlargelanguage}. RAG (Retrieval Augmented Generation) offers a solution to this issue by providing a way to provide external knowledge as context to the models~\cite{NEURIPS2020_6b493230}. Given its practicality, multimodal RAG has also been explored. However, compared to text-only retrieval, it involves retrieving documents comprising of rich information in the form of figures, charts, and tables along with text, increasing complexity and requiring understanding visual and spatial representations~\cite{mei2025surveymultimodalretrievalaugmentedgeneration}. 
Despite these challenges, multimodal RAG has made notable progress with the development of models like ColPali ~\cite{faysse2025colpaliefficientdocumentretrieval}. 
Although representing meaningful progress, these methods exhibit several limitations: %
\emph{(i)} They use pretrained VLMs such as PaliGemma~\cite{beyer2024paligemmaversatile3bvlm} to obtain visual representations; however, these models do not explicitly optimize output visual tokens during pretraining, as the autoregressive loss is applied only to subsequent text tokens, limiting their effectiveness for fine-grained visual retrieval.
\emph{(ii)} They rely heavily on supervised fine-tuning with annotated query–document pairs to achieve cross-modal alignment, which restricts their applicability in domains lacking such labeled data.
\emph{(iii)} Existing methods adopt late-interaction mechanisms such as MaxSim ~\cite{khattab2020colbertefficienteffectivepassage}, originally designed for text retrieval, which assume a one-to-one correspondence between query and document tokens. In visual contexts, this assumption breaks down, as patch-based image tokenization can fragment words or merge multiple words into a single patch, introducing noise in similarity computation during training and ultimately degrading retrieval performance. These shortcomings in the design of existing methods may limit their capabilities for multimodal document retrieval.

\begin{figure*}[t]
    \centering
    \begin{subfigure}[t]{0.22\textwidth}
        \centering
        \includegraphics[width=\linewidth]{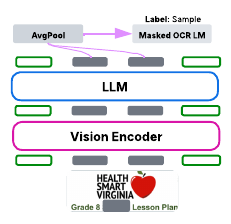}
        \caption{\small Masked OCR LM}
        \label{fig:sub2}
    \end{subfigure}
    \hfill
    \begin{subfigure}[t]{0.38\textwidth}
        \centering
        \includegraphics[width=\linewidth]{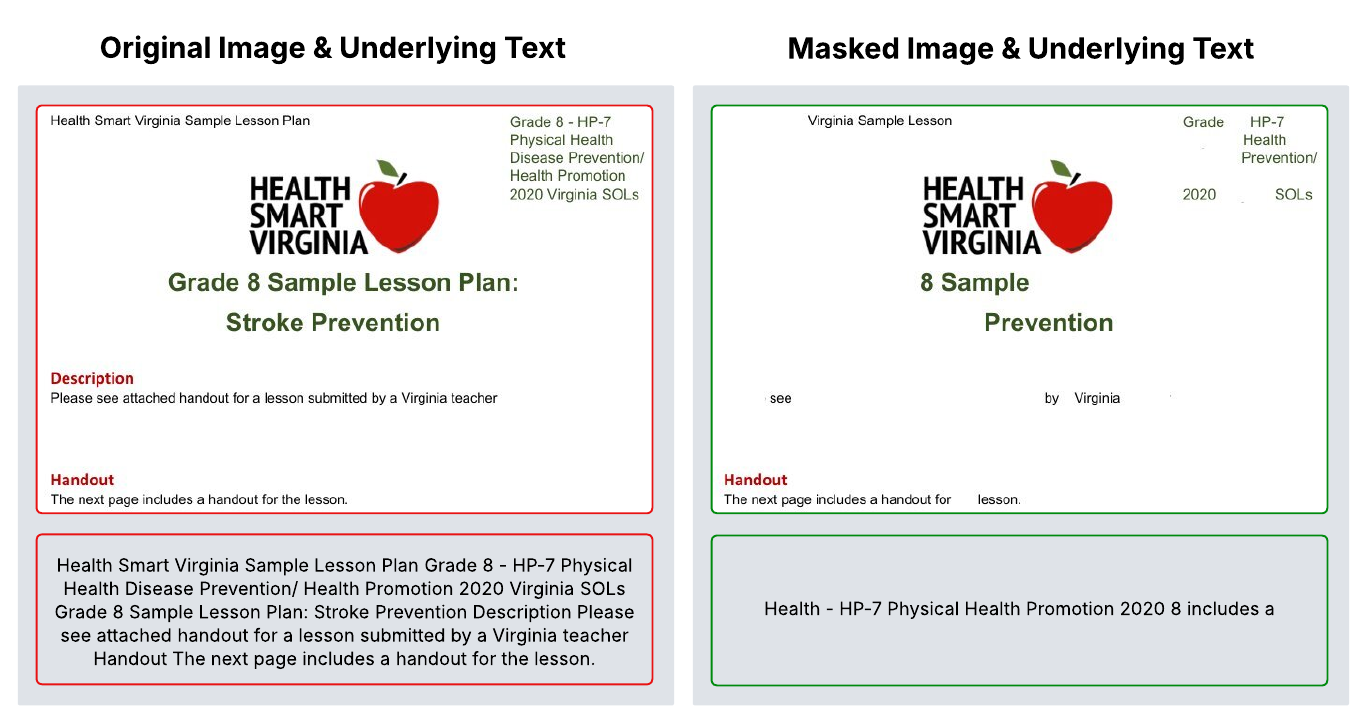}
        \caption{\small Masked Contrastive Learning}
        \label{fig:sub3}
    \end{subfigure}
    \hfill
    \begin{subfigure}[t]{0.30\textwidth}
        \centering
        \includegraphics[width=\linewidth]{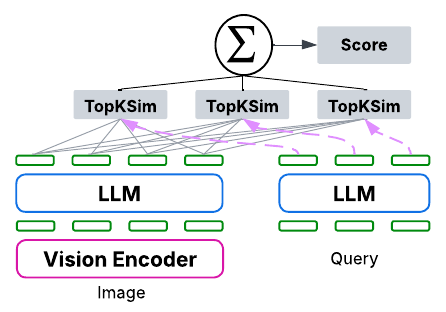}
        \caption{\small Training with TopKSim}
        \label{fig:sub1}
    \end{subfigure}
    \caption{Overview of \colmate{}’s key components. \emph{(a) Masked OCR Language Modeling (MOLM)} explicitly optimizes visual token representations by predicting masked OCR tokens during pretraining. \emph{(b) Masked Contrastive Learning (MaskedCL)} enables self-supervised cross-modal alignment between masked text and document features when query–document pairs are unavailable. \emph{(c) TopKSim} refines the late-interaction mechanism by averaging top-K similarity scores during training to reduce noise from patch-based tokenization in visual documents.}
    \label{fig:main}
\end{figure*}

We address these limitations by introducing \colmate{}, a multimodal document retrieval model designed to capture the rich information embedded in visually dense documents. \colmate{} introduces three complementary components across pretraining, fine-tuning, and late-interaction: \emph{(i)} a masked OCR language modeling objective that explicitly optimizes visual token representations during pretraining, \emph{(ii)} a self-supervised contrastive learning objective that enables cross-modal alignment without relying on annotated query–document pairs, and \emph{(iii)} a refined late-interaction mechanism, TopKSim, that alleviates training noise arising from patch-based tokenization in visual documents. Together, these components bridge the gap between existing retrieval approaches and recent advances in multimodal representation learning, yielding more robust and generalizable multimodal document retrieval.

\colmate{} improves performance over existing methods on both in-domain and out-of-domain ViDoRe benchmarks~\cite{faysse2025colpaliefficientdocumentretrieval, macé2025vidorebenchmarkv2raising}, with particularly notable gains in out-of-domain generalization. The contributions of this work are summarized as follows: (i) we present\colmate{}, a multimodal document retrieval model that integrates three complementary components across pretraining, self-supervised fine-tuning, and late interaction; (ii) we demonstrate consistent performance improvements compared to existing methods across diverse domains; and (iii) we provide detailed ablations and analyses to quantify the contribution of each component. To support future research, we release the model weights publicly at \href{https://huggingface.co/ahmed-masry/ColMate-3B}{https://huggingface.co/ahmed-masry/ColMate-3B}.

\section{Related Work}
\label{sec:related_work}

\subsection{Multimodal Retrieval \& Late Interaction} 
Multimodal retrieval models have significantly advanced through contrastive learning approaches that align visual and textual representations in shared embedding spaces~\cite{radford2021learning, jia2021scaling, zhai2023sigmoid}. These models utilize contrastive learning techniques~\cite{hadsell2006dimensionality, chen2020simple} and are trained on large-scale image-text datasets~\cite{lin2014microsoft, schuhmann2022laion} to enable cross-modal retrieval capabilities, facilitating downstream applications such as visual question answering (VQA)~\cite{liu2023visual, beyer2024paligemmaversatile3bvlm} and multimodal RAG~\cite{chen2022murag, yasunaga2022retrieval}. For efficient dense retrieval, late interaction mechanisms like ColBERT~\cite{khattab2020colbertefficienteffectivepassage} have proven effective by computing fine-grained token-level similarities between queries and documents, and recent works have extended these approaches to multimodal document retrieval, with ColPali~\cite{faysse2025colpaliefficientdocumentretrieval} directly applying ColBERT's MaxSim operation \cite{khattab2020colbertefficienteffectivepassage} to vision-language models like PaliGemma~\cite{beyer2024paligemmaversatile3bvlm}. However, this direct adaptation introduces noise during training because MaxSim assumes a one-to-one correspondence between query and document tokens. This assumption breaks down in visual contexts where patch-based tokenization can fragment words across multiple patches.

\subsection{Visual Document Understanding \& Pretraining}
Visual document understanding focuses on enabling models to comprehend and process documents that often contain complex layouts, diverse fonts, and multimodal content such as images, tables, and charts. Models like DocFormer~\cite{appalaraju2021docformer}, LayoutLMv3~\cite{huang2022layoutlmv3}, and DiT~\cite{li2022dit} integrate textual content with layout and visual information, while approaches like StructTextV2~\cite{yu2023structextv2} and UniDoc~\cite{gu2021unidoc} refine pretraining objectives specifically for document understanding through masked image and language modeling over structured layouts. Recent large-scale efforts like BigDocs~\cite{rodriguez2025bigdocsopendatasettraining} have demonstrated the critical importance of document-specific training data and pretraining for improving document comprehension capabilities. %
Despite these advances, most multimodal document retrieval methods build on general-purpose VLMs like PaliGemma~\cite{beyer2024paligemmaversatile3bvlm}, whose pretraining objectives do not explicitly optimize visual token representations in the final layer. As a result, visual embeddings may not capture the fine-grained document structure and semantics, limiting retrieval effectiveness.

\subsection{Visually Rich Document Retrieval.} 
Retrieving visually rich documents is challenging, as traditional OCR-based text retrieval struggles to capture layout information and visual semantics. Recent models such as DSE~\cite{ma2024unifyingmultimodalretrievaldocument} and VisRAG~\cite{yu2024visrag} leverage vision–language models (VLMs) directly for document retrieval, reducing the need for complex OCR preprocessing pipelines. ColPali~\cite{faysse2025colpaliefficientdocumentretrieval} further extends late-interaction architectures like ColBERT~\cite{khattab2020colbertefficienteffectivepassage} to efficiently match query and document embeddings. Self-supervised objectives such as masked language modeling~\cite{devlin2019bert} and contrastive learning~\cite{hadsell2006dimensionality, chen2020simple} have proven effective for improving representation learning, with extensions like SpanBERT~\cite{joshi2020spanbertimprovingpretrainingrepresenting} and hard-negative contrastive training~\cite{robinson2021contrastivelearninghardnegative} enhancing robustness. However, existing multimodal document retrieval models rely predominantly on supervised fine-tuning with annotated query–document pairs, limiting scalability and generalization to domains where such data is scarce.

\section{Methodology}
\label{colmate}

As presented in \Cref{fig:main},\colmate{} incorporates three novel components: \emph{(i)} MOLM, an OCR-based masked language modeling training objective for improved feature representation, \emph{(ii)} TopKSim, a late-interaction mechanism optimized for vision domain retrieval, and  \emph{(iii)} MaskedCL, a self-supervised contrastive learning objective for scenarios without query-image pairs. 

\subsection{Masked OCR Language Modeling (MOLM)}
ColPali initializes from PaliGemma's pretrained weights \cite{beyer2024paligemmaversatile3bvlm}. However, the pretraining of PaliGemma and similar VLMs often does not directly optimize the vision tokens representations in the last layer, as the autoregressive loss is typically applied only to subsequent text tokens. Consequently, vision features may be suboptimal, hindering retrieval tasks that heavily rely on them.

To address this, we introduce a novel pretraining objective, \emph{Masked OCR Language Modeling (MOLM)}, formulated as follows:

\begin{itemize}
  \item[(1)] Given an input document image, we obtain contextualized visual token embeddings \( V = \{v_1, v_2, \dots, v_N\} \subseteq \mathbb{R}^d \) from the LLM output. Suppose we have an OCR word token \( \{w_j\} \) with its  corresponding bounding box. Let \( B_j \subseteq \{1, \dots, N\} \) denote the indices of visual tokens that spatially overlap with the bounding box of word \( w_j \).
  
  \item[(2)] We randomly select 30\% of the OCR tokens to mask, forming the set \( \mathcal{M} \subset \{w_j\} \). For each masked word \( w_j \in \mathcal{M} \), we compute a pooled visual representation over the tokens in \( B_j \):
  \[
  \bar{v}_j = \frac{1}{|B_j|} \sum_{i \in B_j} v_i
  \]
  
  \item[(3)] Using these pooled embeddings, the model is trained with a masked language modeling objective to predict the masked OCR tokens:
  \[
  \mathcal{L}_{\text{MOLM}} = -\sum_{w_j \in \mathcal{M}} \log p(w_j \mid \bar{v}_j, \theta)
  \]
  where \( p(w_j \mid \bar{v}_j, \theta) \) denotes the probability of predicting the token \( w_j \) from its visual context by the model with parameters \( \theta \).
\end{itemize}

By directly optimizing visual token embeddings through this OCR-based masking objective, MOLM significantly enriches the visual representations, thereby enhancing downstream multimodal retrieval performance.

\subsection{TopKSim}
Current late-interaction methods like MaxSim, originally designed for text-based retrieval models such as ColBERT~\cite{khattab2020colbertefficienteffectivepassage}, are suboptimal when directly applied to document images ~\cite{faysse2025colpaliefficientdocumentretrieval}. This limitation stems from fundamental differences in how text and images are tokenized. In text, tokens usually align well with semantic units such as words. In contrast, visual tokens (derived from image patches) may span multiple words or capture only parts of a word, depending on factors like patch size and font rendering. To address this, we propose \emph{TopKSim}, a novel method that averages the top-$K$ similarity scores during training instead of relying on the single maximum. The hyperparameter $K$ controls the number of top-scoring document tokens considered for each query token. This approach reduces training noise and mitigates the over-reliance on single image patches, acting as a regularizer. This results in more robust retrieval and matching of query-image pairs. 

Formally, given an encoded query $q = \{ q_1, q_2, \dots, q_n \} \subset \mathbb{R}^d$ and an encoded document $d = \{ d_1, d_2, \dots, d_m \} \subset \mathbb{R}^d$, we define the similarity score as:
\[
\text{Score}(q, d) = \sum_{i=1}^{n} \frac{1}{K} \sum_{j \in \mathcal{I}_i} \langle q_i \mid d_j \rangle,
\]
where $\langle q_i \mid d_j \rangle$ denotes the dot product between the $i$-th query token and the $j$-th document token, and $\mathcal{I}_i \subseteq \{1, \dots, m\}$ is the set of indices corresponding to the $K$ largest dot products $\langle q_i \mid d_j \rangle$ for fixed $i$, defined as:

\[
\mathcal{I}_i = \operatorname{arg\,topK}_{j \in \{1, \dots, m\}} \langle q_i \mid d_j \rangle
\]

We employ TopKSim only during training; at inference, we revert to MaxSim, which has shown superior empirical performance.

\subsection{Self-supervised Masked Contrastive Learning (MaskedCL).}
In practical scenarios, there are often abundant PDF documents available, but corresponding annotated query-document pairs may be lacking. To address this scenario, we propose MaskedCL, a self-supervised contrastive learning approach designed to mimic supervised contrastive fine-tuning without relying on labeled queries. MaskedCL integrates contrastive learning principles with a masking-based pretext task specifically tailored for multimodal document retrieval: \emph{(i)} We construct pseudo-queries by randomly masking spans within the text content extracted from PDFs. \emph{(ii)} Correspondingly, we generate masked versions of document images by overlaying white masks on random patches. \emph{(iii)} Finally, we perform contrastive alignment between the masked textual representations and masked visual representations using our TopKSim late interaction mechanism. This strategy encourages robust cross-modal representation learning, even when large portions of textual or visual information are masked. 

Formally, following prior methods~\cite{faysse2025colpaliefficientdocumentretrieval, khattab2020colbertefficienteffectivepassage}, we define the MaskedCL loss using in-batch softmax cross-entropy over positive and hardest negative scores. Given a pseudo-query \( q_k \) and its corresponding masked document image \( d_k \), we compute:
\begin{align*}
s_k^{+} &= \text{Score}(q_k, d_k), \\[4pt]
s_k^{-} &= \max_{l \ne k} \text{Score}(q_k, d_l),
\end{align*}
where \(\text{Score}(q, d)\) denotes the TopKSim score. The MaskedCL loss is then defined as:
\begin{equation*}
\mathcal{L}_{\text{MaskedCL}} = - \frac{1}{b} \sum_{k=1}^{b} \log \left( \frac{ \exp(s_k^{+}) }{ \exp(s_k^{+}) + \exp(s_k^{-}) } \right),
\end{equation*}
with \( b \) representing the batch size.

\definecolor{pali_models}{RGB}{185, 235, 255}
\definecolor{qwen_models}{RGB}{255, 219, 187}

\begin{table*}[t]
\centering
\scriptsize
\begin{tabular}{lrrrrrrrrrrr}
\toprule
& ArxivQ & DocQ & InfoQ & TabF & TATQ & Shift & AI   & Energy & Gov.  & Health & Avg. \\

\midrule
\multicolumn{11}{l}{\textbf{Self-supervised Models}} \\
\rowcolor{pali_models!70} ColMate-Pali-3B (CL) & 58.78 & 41.19 & 76.97 & 69.06 & 46.97 & 67.31 & 91.65 & 88.97 & 88.02 & 87.32 & \cellcolor{pali_models}71.62 \\
\rowcolor{pali_models!70} ColMate-Pali-3B (MaskedCL) \textbf{(\textcolor{blue}{Ours})} & 67.38 & 44.03 & 77.81 & 71.37 & 49.84 & 66.96 & 92.25 & 91.88 & 92.75 & 90.95 & \cellcolor{pali_models}\textbf{74.52} \\

\midrule
\multicolumn{12}{l}{\textbf{Supervised Models}} \\
\rowcolor{qwen_models!70} ColPali-3B    & 83.03 & 58.45 & 85.71 & 87.44 & 70.36 & 77.38 & 97.43 & 95.40 & 96.21 & 96.91 & \cellcolor{qwen_models}84.93 \\
\rowcolor{qwen_models!70} ColPali-3B (Reproduced) & 84.55 & 57.65 & 86.43 & 86.65 & 71.31 & 76.40 & 96.62 & 94.64 & 94.84 & 97.76 & \cellcolor{qwen_models}84.68 \\
\rowcolor{qwen_models!70} ColMate-Pali-3B \textbf{(\textcolor{blue}{Ours})} & 83.68 & 57.52 & 84.15 & 87.65 & 74.06 & 79.84 & 98.36 & 94.15 & 95.34 & 96.65 & \cellcolor{qwen_models}\textbf{85.14} \\

\bottomrule
\end{tabular}
\caption{nDCG@5 scores of\colmate{} and baselines on the ViDoRe V1 (in-domain) benchmark 10 academic and real-world datasets.\colmate{} achieves the highest average performance. In self-supervised settings, MaskedCL outperforms standard contrastive learning (CL).}
\label{tab:results_v1}
\end{table*}

\section{Experimental Setup}
\label{sec:experimental_setup}

\paragraph{Datasets \& Benchmarks}
For masked OCR language modeling (MOLM), we use 4M digital PDF documents from the pdfa-eng-wds dataset\footnote{https://huggingface.co/datasets/pixparse/pdfa-eng-wds}, rendered as images at 96 dpi. From these, 1M documents with their underlying metadata (words and bounding boxes) are used for self-supervised contrastive masked learning. For supervised fine-tuning, we adopt the ViDoRe training split \cite{faysse2025colpaliefficientdocumentretrieval}, which includes 118,695 query-page pairs from synthetic and public datasets.

For evaluation, we use two benchmarks: \emph{(i)} ViDoRe V1 (in-domain) \cite{faysse2025colpaliefficientdocumentretrieval}, which covers 10 academic and real-world datasets (e.g., DocVQA \cite{mathew2021docvqadatasetvqadocument}, InfoVQA \cite{mathew2021infographicvqa}, arXivVQA \cite{arxivvqa}, and real-world domains like energy, government, healthcare, and AI), and \emph{(ii)} ViDoRe V2 (out-of-domain), comprising documents from 9 diverse real-world domains such as biomedical, economics, and ESG. Both benchmarks are multilingual, while training data is exclusively English.

\paragraph{Modeling}
\colmate{} builds upon the ColPali model~\citep{faysse2025colpaliefficientdocumentretrieval}, chosen for its faster runtime. To compare the models efficiency, we measured forward pass times for image encoding on a single H100 GPU using a batch size of 4 and 500 images from the DocVQA benchmark. The average batch times were: ColPali-3B (76 ms) ~\citep{faysse2025colpaliefficientdocumentretrieval}, ColQwen2.5-3B (188 ms) ~\citep{Qwen2-VL}, and ColSmolVLM-256M (100 ms) ~\citep{marafioti2025smolvlmredefiningsmallefficient}. ColPali-3B was the fastest. Notably, ColSmolVLM-256M, despite being 12× smaller than ColPali-3B, was slower, primarily due to differences in image preprocessing. ColPali-3B processes images at 448×448 resolution, while ColSmolVLM-256M uses a 2048×2048 resolution split into 17 crops of 512×512. Given its superior speed, we selected ColPali-3B for our\colmate{} framework. We initialize our training from the \emph{paligemma-448-base} checkpoint.

\paragraph{Hyperparameters}
For MOLM, we train on the 4M pages from PDFA for 1 epoch, with a learning rate of 3e-5 and batch size 64 using Adam~\cite{kingma2017adammethodstochasticoptimization}.
For MaskedCL, we train on 1M documents for one epoch with a learning rate of 2e-5 and batch size 256 using paged AdamW (8-bit). We use LoRA~\cite{hu2021loralowrankadaptationlarge} with $\alpha=32$ and rank $r=32$ for the LLM layers and the projection layer.
For effective masking, we adopt the SpanBERT masking mechanism~\cite{joshi2020spanbertimprovingpretrainingrepresenting}, masking contiguous spans sampled from a geometric distribution with a maximum length of 10 and probability $p=0.2$. To simulate real-world queries referencing small document sections, we apply an aggressive 80\% masking probability to text. To avoid trivial text-to-image matching, we also randomly mask 50\% of OCR words in the images with white masks. Finally, in the full supervised finetuning setup, we pre-train\colmate{} using MaskedCL before the full supervised fine-tuning.

For supervised fine-tuning, we follow ColPali v1.3 settings\footnote{https://huggingface.co/vidore/colpali-v1.3}, using a learning rate of 5e-5, batch size 256, 1000 warmup steps, and 3 training epochs of the ViDoRe training set ~\citep{faysse2025colpaliefficientdocumentretrieval}.

\noindent Finally, we set \( K = 5 \) for TopKSim, as it provided the best performance in preliminary experiments comparing \( K \in \{3, 5, 10\} \). Experiments were conducted on machines equipped with 8×H100 and 4×H100 GPUs.

\begin{table*}[t]
\centering
\scriptsize
\scalebox{0.88}{
\begin{tabular}{lrrrrrrrrrr}
\toprule
& MIT Bio. & Econ. Mac. & ESG & AXA & ESG-M & AXA-M & MIT Bio.-M & Econ. Mac-M & ESG Human & Avg. \\
\midrule

\midrule
\multicolumn{11}{l}{\textbf{Self-supervised Models}} \\
\rowcolor{pali_models!70} ColMate-Pali-3B (CL) & 42.18 & 47.16 & 20.89 & 46.64 & 31.90 & 34.76 & 35.68 & 38.86 & 44.95 & \cellcolor{pali_models}38.11  \\
\rowcolor{pali_models!70} ColMate-Pali-3B (MaskedCL) \textbf{(\textcolor{blue}{Ours})} & 50.18 & 47.21 & 24.53 & 40.24 & 35.36 & 36.61 & 44.61 & 40.22 & 54.51 & \cellcolor{pali_models}\textbf{41.50} \\

\midrule
\multicolumn{11}{l}{\textbf{Supervised Models}} \\
\rowcolor{qwen_models!70} ColPali-3B    & 59.70 & 51.60 & 57.00 & 59.80 & 55.70 & 50.10 & 56.50 & 49.90 & 51.10 & \cellcolor{qwen_models}54.60 \\
\rowcolor{qwen_models!70} ColPali-3B (Reproduced)   & 60.20 & 54.01 & 52.14 & 54.61 & 53.13 & 45.34 & 58.43 & 49.01 & 59.17 & \cellcolor{qwen_models}54.00 \\
\rowcolor{qwen_models!70} ColMate-Pali-3B \textbf{(\textcolor{blue}{Ours})} & 60.99 & 55.99 & 54.15 & 67.01 & 53.44 & 50.61 & 59.31 & 54.14 & 62.82 & \cellcolor{qwen_models}\textbf{57.61} \\

\bottomrule
\end{tabular}
}
\caption{nDCG@5 scores on ViDoRe V2 (out-of-domain) across 9 real-world domains.\colmate{} achieves the best average and demonstrates stronger generalization.}
\label{tab:results_v2}
\end{table*}

\section{Evaluation}

\subsection{Main Results}
We present results on ViDoRe V1 (in-domain) and ViDoRe V2 (out-of-domain) in Tables \ref{tab:results_v1} and \ref{tab:results_v2}.\colmate{} outperforms the ColPali baseline on both benchmarks. On ViDoRe V1, ColMate-Pali-3B achieves an average nDCG@5 of 85.14, surpassing both the original ColPali-3B (84.93) and our reproduction (84.68). On ViDoRe V2,\colmate{} reaches 57.61, compared to 54.60 for ColPali-3B and 54.00 for our reproduction, demonstrating strong generalization to unseen domains.

To simulate scenarios without image-query pairs, we evaluate self-supervised Masked Contrastive Learning (MaskedCL) against standard Contrastive Learning (CL) without masking. MaskedCL, even without supervised finetuning, achieves competitive performance—74.52 on ViDoRe V1 and 41.50 on ViDoRe V2—highlighting its effectiveness in low-resource settings. The performance gap with supervised finetuning is especially narrow on subsets such as AI, Energy, Gov., and Health, which represent real-world documents. MaskedCL also outperforms standard CL by 2.90 and 3.39 on V1 and V2, respectively, making it a strong choice when training data lacks image-query pairs.

\subsection{Ablation Studies}
\begin{figure}
    \centering
    \includegraphics[width=0.99\linewidth
    ]{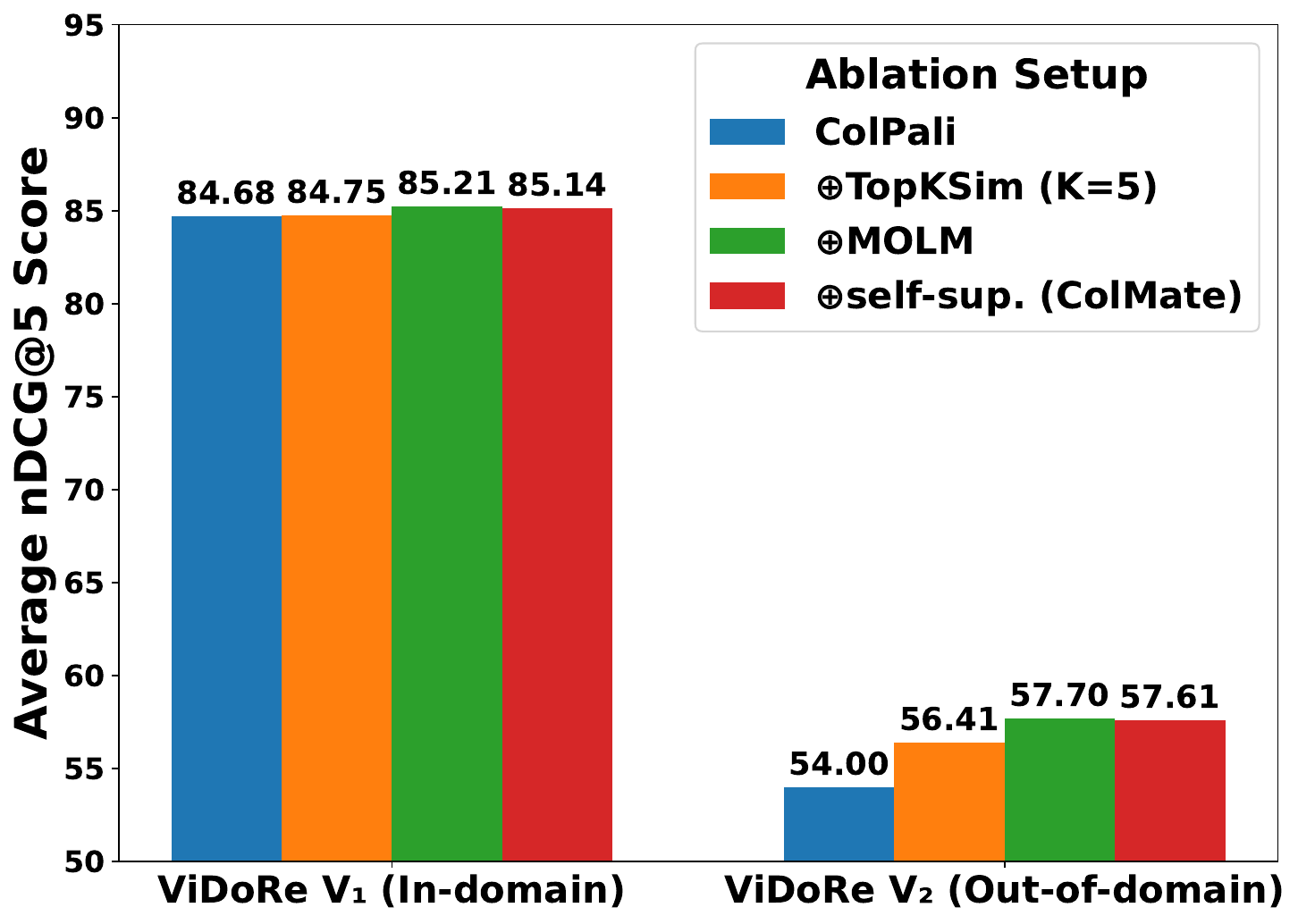}
    \caption{\textbf{Ablation Studies:} Impact of\colmate{} Components on ViDoRe Benchmarks (Average nDCG@5)}
    \label{fig:ablations}
\end{figure}

To better understand the individual contributions of each proposed component of ColMate, we perform a detailed ablation study using the PaliGemma-3B model ~\citep{beyer2024paligemmaversatile3bvlm}. The effectiveness of each component is measured using the ViDoRe V1 and ViDoRe V2 benchmarks, and results are reported using the nDCG@5 metric in Figure \ref{fig:ablations} and more detailed numbers on Table \ref{tab:ablations_combined}.

\paragraph{TopKSim vs MaxSim}
We compare our TopKSim mechanism with the MaxSim baseline \cite{khattab2020colbertefficienteffectivepassage, faysse2025colpaliefficientdocumentretrieval} for multimodal document retrieval. Two ColPali models are fine-tuned from \emph{paligemma-448-base} using identical hyperparameters from Section \ref{sec:experimental_setup}, one with TopKSim (K=5), the other with MaxSim.

As shown in Figure \ref{fig:ablations}, TopKSim achieves better generalization on the out-of-domain ViDoRe V2 benchmark (56.41 vs. 54.00) and performs slightly better on the in-domain ViDoRe V1. The improvement is especially notable on the TabFQuAD subset (see Table ~\ref{tab:ablations_combined}), which lacks similar examples in the training set, further highlighting the improved robustness and generalization offered by TopKSim.

\paragraph{Adding Masked OCR Modeling}
We assess the impact of applying Masked OCR Language Modeling (MOLM) before supervised finetuning and observe consistent performance gains across both benchmarks. On the out-of-domain ViDoRe V2, MOLM increases average score from 56.41 to 57.70, supporting our hypothesis that modeling OCR-masked tokens improves vision-language representations for visual documents.

\paragraph{Adding Self-Supervised Finetuning}
We evaluate the impact of applying our self-supervised Masked Contrastive Learning (MaskedCL) framework before supervised finetuning. As shown in Figure \ref{fig:ablations}, MaskedCL offers no performance gains when supervised data is abundant, suggesting its limited utility in data-rich scenarios. However, Tables \ref{tab:results_v1} and \ref{tab:results_v2} illustrate that MaskedCL alone achieves performance comparable to supervised finetuning across several ViDoRe subsets, highlighting its effectiveness in low-resource contexts.

\subsection{Experiments with More Powerful Backbone Models}

To examine the scalability of our framework, we applied\colmate{} to a more capable backbone, Qwen2.5-VL-3B~\citep{Qwen2-VL}, following the same pretraining and fine-tuning procedure used for ColPali.
Table \ref{tab:qwen_results} reports performance on ViDoRe V1 (in-domain) and ViDoRe V2 (out-of-domain) benchmarks.
\colmate{} improves the out-of-domain performance of Qwen2.5-VL-3B by 0.86 nDCG@5, indicating that the inductive bias introduced by\colmate{} has a smaller effect as backbone models become stronger and benefit from richer pretraining data.
Nevertheless, the framework delivers larger gains with ColPali, which is substantially faster (86 ms vs. 188 ms for ColQwen2.5-VL), translating into tangible benefits in resource-constrained or latency-sensitive settings.

\section{Conclusion}
We introduce \colmate{}, a multimodal document retrieval model that addresses key limitations of existing methods by combining three simple but effective ideas: masked OCR language modeling for richer vision-text representations, masked contrastive learning for effective self-supervised alignment, and TopKSim for robust similarity aggregation during training. %
\colmate{} consistently improves retrieval performance over existing methods on both in-domain and out-of-domain ViDoRe benchmarks, with particularly strong generalization to unseen domains. Our extensive ablation studies show the individual contribution of each component, validating our design choices. Our results highlight \colmate{}’s utility for practical retrieval scenarios, particularly where annotated image-query pairs are scarce. 

\noindent As future work, we plan to extend \colmate{} to additional model architectures and evaluate it on broader multimodal document retrieval benchmarks.

\section*{Limitations}
This work applies the\colmate{} framework primarily to the PaliGemma model ~\citep{beyer2024paligemmaversatile3bvlm} for computational efficiency. Future work will extend\colmate{} to other models such as SmolVLM~\cite{marafioti2025smolvlmredefiningsmallefficient}. While\colmate{} provides only modest gains on in-domain benchmarks (Table~\ref{tab:results_v1}), it delivers substantial improvements on out-of-domain tasks (Table~\ref{tab:results_v2}). TopKSim also introduces an additional hyperparameter ($K$) that may require tuning. Lastly, MaskedCL offers limited benefits when annotated query-image pairs are abundant, but is highly effective as a self-supervised method when such data is unavailable.

\section*{Ethical Considerations}
We complied with the terms of use and licenses for the ViDoRe benchmarks and the PaliGemma model, which were used solely for research puroses in our work. Our models are not generative; they encode documents and queries for retrieval tasks. Therefore, we do not anticipate risks typically associated with large language models, such as hallucinations or harmful content generation. Still, proper evaluation is necessary before deploying these models in real-world scenarios. AI writing tools were used only to enhance the paper’s writing.

\bibliography{custom}

\appendix

\begin{table*}[h]
  \centering
  \scriptsize

  \resizebox{\textwidth}{!}{%
    \begin{tabular}{lrrrrrrrrrrr}
      \toprule
      \multicolumn{12}{c}{\textbf{ViDoRe V\textsubscript{1} (nDCG@5) (In-domain)}} \\
      \cmidrule(lr){1-12}
      Method               & ArxivQ & DocQ & InfoQ & TabF & TATQ & Shift & AI  & Energy & Gov. & Health & Avg.  \\
      \midrule
      ColPali                & 84.55 & 57.65 & 86.43 & 86.65 & 71.31 & 76.40 & 96.62 & 94.64 & 94.84 & 97.76   & 84.68  \\
      \quad \(\oplus\) TopKSim (K=5)      & 83.08 & 57.40 & 85.11 & 90.43 & 71.02 & 76.41 & 97.63 & 94.66 & 94.84 & 96.94 & 84.75   \\
      \quad \quad \(\oplus\) MOLM   & 83.80 & 55.98 & 84.68 & 88.60 & 74.79 & 80.51 & 98.01 & 94.95 & 94.93 & 95.89  & 85.21  \\
      \quad \quad \quad \(\oplus\) self-sup. (ColMate)      & 83.68 & 57.52 & 84.15 & 87.65 & 74.06 & 79.84 & 98.36 & 94.15 & 95.34 & 96.65 & 85.14  \\
      \bottomrule
    \end{tabular}%
  }
  \vspace{1em}
  \resizebox{\textwidth}{!}{%
    \begin{tabular}{lrrrrrrrrrr}
      \toprule
      \multicolumn{11}{c}{\textbf{ViDoRe V\textsubscript{2} (nDCG@5) (Out-of-domain)}} \\
      \cmidrule(lr){1-11}
      Method               & MIT Bio. & Econ. Mac. & ESG   & AXA   & ESG–M & AXA–M & Bio.–M & Econ.–M & ESG Human & Avg.   \\
      \midrule
      ColPali (Reproduced)             & 60.20 & 54.01 & 52.14 & 54.61 & 53.13 & 45.34 & 58.43 & 49.01 & 59.17 & 54.00 \\
      \quad \(\oplus\) TopKSim (K=5)    & 60.53 & 54.46 & 55.87 & 61.13 & 55.91 & 47.75 & 57.99 & 52.34 & 61.71 & 56.41 \\
      \quad \quad \(\oplus\) MOLM                 & 60.55 & 57.64 & 58.50 & 54.08 & 57.84 & 50.16 & 58.28 & 57.00 & 65.22 & 57.70 \\
      \quad \quad \quad \(\oplus\) self-sup. (ColMate)      & 60.99 & 55.99 & 54.15 & 67.01 & 53.44 & 50.61 & 59.31 & 54.14 & 62.82 & 57.61 \\
      \bottomrule
    \end{tabular}%
  }
\caption{Ablation studies showing the impact of the different\colmate{} components on the ViDoRe V\textsubscript{1} and V\textsubscript{2} benchmarks (nDCG@5).}
  \label{tab:ablations_combined}

\end{table*}

\section{Appendices}

\subsection{Extended and Detailed Results}
We present detailed ablation results in Table \ref{tab:ablations_combined}.

\begin{table*}[h]
  \centering
  \scriptsize

  \resizebox{\textwidth}{!}{%
    \begin{tabular}{lrrrrrrrrrrr}
      \toprule
      \multicolumn{12}{c}{\textbf{ViDoRe V\textsubscript{1} (nDCG@5) (In-domain)}} \\
      \cmidrule(lr){1-12}
      Method               & ArxivQ & DocQ & InfoQ & TabF & TATQ & Shift & AI  & Energy & Gov. & Health & Avg.  \\
      \midrule
      ColQwen2.5-VL-3B & 89.22 & 63.22 & 92.37 & 91.12 & 81.10 & 87.30 & 99.63 & 95.89 & 96.41 & 97.89 & 89.41 \\
    \rowcolor{gray!10}
    \textbf{ColMate-Qwen2.5-VL-3B} & 90.24 & 61.10 & 93.68 & 91.51 & 81.88 & 90.24 & 99.26 & 96.39 & 96.54 & 98.13 & \textbf{89.89} \\
      \bottomrule
    \end{tabular}%
  }
  \vspace{1em}
  \resizebox{\textwidth}{!}{%
    \begin{tabular}{lrrrrrrrrrr}
      \toprule
      \multicolumn{11}{c}{\textbf{ViDoRe V\textsubscript{2} (nDCG@5) (Out-of-domain)}} \\
      \cmidrule(lr){1-11}
      Method               & MIT Bio. & Econ. Mac. & ESG   & AXA   & ESG–M & AXA–M & Bio.–M & Econ.–M & ESG Human & Avg.   \\
      \midrule
      ColQwen2.5-VL-3B & 63.64 & 59.78 & 57.39 & 60.29 & 57.38 & 53.19 & 61.12 & 56.52 & 68.39 & 59.74 \\
        \rowcolor{gray!10}
        \textbf{ColMate-Qwen2.5-VL-3B} & 62.06 & 59.55 & 60.06 & 65.82 & 60.21 & 56.39 & 60.33 & 52.05 & 68.94 & \textbf{60.60} \\
      \bottomrule
    \end{tabular}%
  }
\caption{nDCG@5 results for applying \colmate{} to Qwen2.5-VL-3B.}
  \label{tab:qwen_results}

\end{table*}

\end{document}